\relax

\documentclass[letterpaper]{article}
\usepackage{times}
\usepackage{helvet}
\usepackage{courier}
\usepackage[hyphens]{url}
\usepackage{graphicx}
\usepackage{subcaption}
\usepackage{multirow}
\usepackage{pgf}

\usepackage{todonotes}
\usepackage{xspace}
\usepackage{amsmath,amsthm,amssymb}
\usepackage{thmtools,mathtools}

\usepackage{booktabs}

\usepackage{algorithm}
\usepackage{algorithmicx}
\usepackage[noend]{algpseudocode}
\algrenewcommand\alglinenumber[1]{\tiny #1:}

\usepackage[noabbrev]{cleveref}

\usepackage{tikz}
\usetikzlibrary{positioning, shapes, calc}

\usepackage{enumitem}
\setitemize{noitemsep,topsep=0pt,parsep=0pt,partopsep=0pt,leftmargin=0pt,itemindent=10pt}
\setenumerate{noitemsep,topsep=0pt,parsep=0pt,partopsep=0pt,leftmargin=0pt,itemindent=10pt}

\newcommand{\myparagraph}[1]{\vskip 4pt \noindent \textbf{#1. }}

\newcommand{\tuple}[1]{\langle #1 \rangle}
\newcommand{\eqbydef}{\: \dot{=} \:}

\newcommand{\prationals}{\mathbb{Q}_{>=0}\xspace}
\newcommand{\sprationals}{\mathbb{Q}_{>0}\xspace}

\newcommand{\rstart}{\textsc{Start}\xspace}
\newcommand{\rend}{\textsc{End}\xspace}

\newtheorem{theorem}{Theorem}[section]
\newtheorem{definition}{Definition}[section]
\newcommand{\deftitle}[1]{\textbf{#1}\xspace}

\newcommand{\planstart}{\pi^\vdash\xspace}
\newcommand{\planend}{\pi^\dashv\xspace}
\newcommand{\actstart}{\mathit{act}^\vdash\xspace}
\newcommand{\actend}{\mathit{act}^\dashv\xspace}
\newcommand{\condstart}{\mathit{cnd}^\vdash\xspace}
\newcommand{\condend}{\mathit{cnd}^\dashv\xspace}
\newcommand{\condinst}{\mathit{cnd}^{[]}\xspace}
\newcommand{\effadd}{\mathit{eff}^+\xspace}
\newcommand{\effdel}{\mathit{eff}^-\xspace}

\newcommand{\sstodo}{\lambda\xspace}
\newcommand{\sstn}{\chi\xspace}
\newcommand{\sslast}{\omega\xspace}

\newcommand{\mktp}[2]{\tuple{#1, #2}}
\newcommand{\mkptp}[1]{\tuple{#1}}
\newcommand{\mkidtp}[3]{\tuple{#2, #3, #1}}

\newcommand{\hsp}{\ensuremath{\textsc{HSP}}\xspace}
\newcommand{\painter}{\ensuremath{\textsc{Painter}}\xspace}
\newcommand{\Driverlog}{\ensuremath{\textsc{Driverlog}}\xspace}
\newcommand{\Satellite}{\ensuremath{\textsc{Satellite}}\xspace}
\newcommand{\TMS}{\ensuremath{\textsc{TMS}}\xspace}
\newcommand{\FloorTile}{\ensuremath{\textsc{Floortile}}\xspace}
\newcommand{\MAP}{\ensuremath{\textsc{MapAnalyser}}\xspace}
\newcommand{\Matchcellar}{\ensuremath{\textsc{MatchCellar}}\xspace}
\newcommand{\majsp}{\ensuremath{\textsc{MAJSP}}\xspace}

\newcommand{\optic}{\textsc{Optic}\xspace}
\newcommand{\fape}{\textsc{Fape}\xspace}
\newcommand{\tpack}{\textsc{Tpack}\xspace}
\newcommand{\itsat}{\textsc{ITSAT}\xspace}
\newcommand{\anmlsmt}{\textsc{Anml smt}\xspace}
\newcommand{\tamer}{\textsc{Tamer}\xspace}
\newcommand{\comptamer}{\textsc{Comp. Tamer}\xspace}

\newcommand{\spin}{\textsc{Spin}\xspace}
\newcommand{\lcp}{\textsc{LCP}\xspace}

\newcommand{\tpackcol}{\parbox[t]{0.7cm}{\linespread{.7}\selectfont\centering\scriptsize \tpack}}
\newcommand{\fapecol}{\parbox[t]{0.7cm}{\linespread{.7}\selectfont\centering\scriptsize \fape}}
\newcommand{\anmlsmtcol}{\parbox[t]{0.65cm}{\linespread{.6}\selectfont\centering\scriptsize \anmlsmt}}
\newcommand{\tamercol}{\parbox[t]{0.7cm}{\linespread{.7}\selectfont\centering\scriptsize \tamer}}

\newcommand{\opticclipcol}{\parbox[t]{0.65cm}{\linespread{.7}\selectfont\centering\scriptsize \optic\\clip}}
\newcommand{\opticcontainercol}{\parbox[t]{0.65cm}{\linespread{.7}\selectfont\centering\scriptsize \optic\\cont.}}
\newcommand{\itsatclipcol}{\parbox[t]{0.72cm}{\linespread{.7}\selectfont\centering\scriptsize \itsat\\clip*}}
\newcommand{\itsatcontainercol}{\parbox[t]{0.72cm}{\linespread{.7}\selectfont\centering\scriptsize \itsat\\cont.*}}
\newcommand{\opticcol}{\parbox[t]{0.65cm}{\linespread{.7}\selectfont\centering\scriptsize \optic}}
\newcommand{\itsatcol}{\parbox[t]{0.72cm}{\linespread{.7}\selectfont\centering\scriptsize \itsat}}

\title{Temporal Planning with Intermediate Conditions and Effects}
\author{Alessandro Valentini, Andrea Micheli and Alessandro Cimatti\\\small{Fondazione Bruno Kessler, Italy}\\\small{\{alvalentini, amicheli, cimatti\}@fbk.eu}}
\date{}

\begin{document}

\maketitle


\begin{abstract}
Automated temporal planning is the technology of choice when controlling systems that can execute more actions in parallel and when temporal constraints, such as deadlines, are needed in the model. One limitation of several action-based planning systems is that actions are modeled as intervals having conditions and effects only at the extremes and as invariants, but no conditions nor effects can be specified at arbitrary points or sub-intervals.

In this paper, we address this limitation by providing an effective heuristic-search technique for temporal planning, allowing the definition of actions with conditions and effects at any arbitrary time within the action duration. We experimentally demonstrate that our approach is far better than standard encodings in PDDL 2.1 and is competitive with other approaches that can (directly or indirectly) represent intermediate action conditions or effects.
\end{abstract}

\section{Introduction}
\label{sec:intro}

Automated temporal planning concerns the synthesis of strategies to
reach a desired goal with a system that is formally specified by
providing an initial condition together with the possible actions that
can drive it in presence of temporal constraints.
In this context, actions become intervals (instead of being
instantaneous as in classical planning) that have a duration (possibly
subject to metric constraints). Similarly, plans are no longer simple
sequences of actions, but they are schedules. Automated temporal
planning received considerable attention in the literature, and the
definition of the standard PDDL 2.1 language \cite{pddl21} fueled the
research of effective search-based techniques to solve the problem
\cite{popf,tfd,itsat}.

One limitation of many approaches encountered in several practical
applications is that conditions and effects of each action can be only
defined when the action starts or terminates, or as an invariant
condition over all the action duration.
This has been recognized \cite{container-actions,micheli-aij} as one
of the major limitations of the PDDL 2.1 language, even if some
compilation-based approaches are known. For example, this is crucial
for the modeling of deadlines that are introduced as a consequence of
an action.

Relaxing this limitation means allowing Intermediate Conditions and
Effects (ICE): hence permitting the definition of actions with
conditions being checked at arbitrary (possibly punctual)
sub-intervals within the action duration and with effects happening at
arbitrary points within the action interval.
%
%
Notably, the ANML \cite{anml} planning formalism offers this feature,
but few planners support this. Our goal in this paper is to natively
offer effective support for planning in domains with ICE.
%

The contributions of this paper are twofold.
First, we present a heuristic-search planning technique able to solve
temporal planning problems with ICE.  Planners based on heuristic
search techniques are currently the state of the art in several areas
of planning; however, none is supporting ICE. We fill this gap by
defining a suitable search space, generalizing the lifted-time approach
of POPF \cite{popf} and providing a powerful relaxation
of the problem for automatic heuristic computation.
Second, we present an automated code-generation technique that produces a
domain-dependent planner from a given model of the system.
In particular, we define a method that maintains the generality of a
model-based, domain independent technique, while providing the
computational advantages of domain-dependent implementations.
We read a planning instance with ICE and generate an executable
embedding our search technique, specialized for the characteristics of
the instance. The resulting executable is very time-efficient and can
solve a range of different problems without recompilation.

We experimentally evaluate the proposed technique by implementing it
in a planner called \tamer and comparing against state of the art
tools. The comparison comprises domains from the literature and
domains inspired by industrial projects where time and temporal
constraints are key aspects and the use of ICE facilitates the
modeling.  Our results show that our technique, thanks to the native
support for ICE, is significantly faster and is able to solve many
more problems than the state of the art tools on the
industrially-inspired domains.
We also experimentally evaluated the domain-dependent planner
generation: our analysis shows that the approach scales even
better than \tamer.

\section{Problem Definition}
\label{sec:problem}

Syntactically, we define a planning problem with ICE analogously to
usual action-based temporal planning problems -- e.g. \cite{pddl21} --
but we allow for conditions and effects to be specified at times
relative to the start or the end of the action instance they belong
to.

\begin{definition}
  An \deftitle{ICE effect} on predicate $p$ at relative time $\tau$ is a tuple $\tuple{\tau, p}$ where $\tau$ is either $\rstart + k$ or $\rend - k$ with $k \in \prationals$.
  An \deftitle{ICE condition}\footnote{We only formalize closed condition intervals; open and semi-open intervals are supported by our implementation.} on predicate $p$ in the relative interval $[\tau_1, \tau_2]$ is a tuple $\tuple{[\tau_1, \tau_2], p}$ where $\tau_i$ is either $\rstart + k_i$ or $\rend - k_i$ with $k_i \in \prationals$.
\end{definition}

Intuitively, an ICE effect (that can be an add or a delete) is applied
at the time specified by $\tau$ that is relative to the start or the
ending time of the action this effect belongs to. Similarly,
conditions are checked on the closed intervals with extremes relative
to the interval extremes of the action they belong to.
In addition, we support timed-initial-literals (TILs) \cite{pddl3}
expressed as a set of ICE effects where $\rstart$ refers to the
beginning of time (i.e. to time $0$) and $\rend$ indicates the end of
the plan execution (i.e the makespan). Similarly, we allow for timed goals (both
instantaneous and durative) as a set of ICE conditions with the same
interpretation of $\rstart$ and $\rend$.

\begin{definition}
  A \deftitle{problem with ICE}
  is a tuple $\tuple{P, A, I, T, G}$ where
  $P$ is a finite set of Boolean predicates;
  $A$ is a set of actions, each action $a$ has a minimal ($d^{min}_a$) and maximal $d^{max}_a$ duration, a set of ICE conditions $C_a$, a set of add ICE effects $E^+_a$ and a set of delete ICE effects $E^-_a$ (with $E^+_a \cap E^-_a = \emptyset$);
  $I \subseteq P$ is the initial state;
  $T$ is a set of ICE effects partitioned in add ($T^+$) and delete ($T^-$), representing TILs; and
  $G$ is a set of ICE goal conditions.
\end{definition}

A plan for an ICE planning problem is analogous to a plan for a
standard temporal planning problem: we have a finite set of action
instances to be executed, each having a specified starting time $t$
and a specified duration $d$.

\begin{definition}
  A \deftitle{plan} $\pi$ is a finite set of tuples $\tuple{t, a, d}$ where
  $t \in \prationals$;
  $a \in A$; and
  $d \in \sprationals$ with $d^{min}_a \le d \le d^{max}_a$.
  The makespan is $ms_\pi \eqbydef max(\{t+d \mid \tuple{t, a, d} \in \pi\})$.
\end{definition}

Semantically, a plan is valid if the execution of the plan respects
all the action conditions and goals and if no two contradicting
effects are applied concurrently.

\begin{definition}
  The set of add (resp. delete) effects of a plan $\pi$ for a planning
  problem with ICE $\mathcal{P} \eqbydef \tuple{P, A, I, T, G}$ is the
  set $E^+_{\pi,\mathcal{P}}$ (resp. $E^-_{\pi,\mathcal{P}}$) defined as the union of the following sets:
  {\small
  \begin{itemize}
    \item $\{\tuple{t + k, p} \mid \tuple{t, a, d} \in \pi, \tuple{\rstart + k, p} \in E^+_a\}$ (resp $\in E^-_a$);
    \item $\{\tuple{t + d - k, p} \mid \tuple{t, a, d} \in \pi, \tuple{\rend - k, p} \in E^+_a\}$ (resp $\in E^-_a$);
    \item $\{\tuple{k, p} \mid \tuple{\rstart + k, p} \in T^+\}$ (resp $\in T^-$);
    \item $\{\tuple{ms_\pi - k, p} \mid \tuple{\rend - k, p} \in T^+\}$ (resp $\in T^-$).
  \end{itemize}
  }

  \noindent
  We define $E_{\pi,\mathcal{P}}$ as $E^+_{\pi,\mathcal{P}} \cup E^-_{\pi,\mathcal{P}}$.
\end{definition}
\noindent
Intuitively, we assign a time to all the effects of the plan
actions and the TILs. Similarly, we can collect all the conditions
imposed by either the plan actions or the timed goals.

\begin{definition}
  The set of conditions of a plan $\pi$ for a planning
  problem with ICE $\mathcal{P} \eqbydef \tuple{P, A, I, T, G}$ is the
  set $C_{\pi,\mathcal{P}}$ defined as the union of the following sets:
  {\small
    \begin{itemize}
    \item $\{ \tuple{[t + k_1, t + k_2], p} \mid \tuple{t, a, d} \in \pi, \tuple{[\rstart + k_1, \rstart + k_2], p} \in C_a \} ;$
    \item $\{\tuple{[t + k_1 , t + d - k_2], p} \mid \tuple{t, a, d} \in \pi, \tuple{[\rstart + k_1, \rend - k_2], p} \in C_a \} ;$
    \item $\{\tuple{[t + d - k_1 , t + k_2], p} \mid \tuple{t, a, d} \in \pi, \tuple{[\rend - k_1, \rstart + k_2], p} \in C_a \} ;$
    \item $\{\tuple{[t + d - k_1 , t + d - k_2], p} \mid \tuple{t, a, d} \in \pi, \tuple{[\rend - k_1, \rend - k_2], p} \in C_a \} ;$

    \item $\{\tuple{[k_1, k_2], p} \mid \tuple{[\rstart + k_1, \rstart + k_2], p} \in G\} ;$
    \item $\{\tuple{[k_1, ms_\pi - k_2], p} \mid \tuple{[\rstart + k_1, \rend - k_2], p} \in G\} ;$
    \item $\{\tuple{[ms_\pi - k_1, k_2], p} \mid \tuple{[\rend - k_1, \rstart + k_2], p} \in G\} ;$
    \item $\{\tuple{[ms_\pi - k_1, ms_\pi - k_2], p} \mid \tuple{[\rend - k_1, \rend - k_2], p} \in G\}$.
    \end{itemize}
  }
\end{definition}

We can now define the semantics of the ICE problem by first explaining
what is a trace induced by a plan and then imposing the validity
conditions on such a trace.

\begin{definition}
  A \deftitle{trace} of $\mathcal{P} \eqbydef \tuple{P, A, I,
    T, G}$ for a plan $\pi$ and a predicate $p \in P$ is a function
  $V_p : \prationals \rightarrow \mathbb{B}$, assigning a Boolean
  value to $p$ at each time:
  {\small
    \begin{itemize}
    \item $V_p(0) \eqbydef \top \mbox{ if $p \in I$}$; \ \ $V_p(0) \eqbydef \bot \mbox{ if $p \not \in I$}$;
    \item $V_p(t) \eqbydef V_p(0) \mbox{ if $t \le min(\{w \mid \tuple{w, p} \in E_{\pi,\mathcal{P}}\})$}$;
    \item $V_p(t) \eqbydef \top \mbox{ if } \tuple{max(\{w \mid w < t, \tuple{w, p} \in E_{\pi,\mathcal{P}}\}), p} \in E^+_{\pi,\mathcal{P}}$;
    \item $V_p(t) \eqbydef \bot \mbox{ if } \tuple{max(\{w \mid w < t, \tuple{w, p} \in E_{\pi,\mathcal{P}}\}), p} \in E^-_{\pi,\mathcal{P}}$.
    \end{itemize}
  }
\end{definition}

\begin{definition}
  A \deftitle{plan $\pi$ is valid} for
  $\mathcal{P} \eqbydef \tuple{P, A, I, T, G}$ if
  $E^+_{\pi,\mathcal{P}} \cap E^-_{\pi,\mathcal{P}} = \emptyset$ and
  for each condition $\tuple{[t_1, t_2], p} \in C_{\pi,\mathcal{P}}$,
  $V_p(k) = \top$ for all $t_1 \le k \le t_2$.
\end{definition}



\section{Heuristic Search for ICE}
\label{sec:search}

We can now present our heuristic-search method for solving planning problems
with ICE by first presenting the search-space design and then a
relaxation used to compute heuristic values.
In the following, we assume a temporal planning problem with ICE
$\mathcal{P} \eqbydef \tuple{P, A, I, T, G}$ is given.

The general idea behind the engineering of our search-space is to
maintain the temporal information symbolic while using explicit
``propositional'' states, similarly to planners such as
POPF~\cite{popf}. Ideally, we separately consider effects and
condition startings/endings as atomic ``events'' that change the
state of the search. We encode such events as \emph{time-points}.

\begin{definition}
  \label{def:time-point}
  A \textbf{time-point} is either:
  $\tuple{\planstart}$, indicating the start of plan instant;
  $\tuple{\planend}$, indicating the end of plan instant;
  $\tuple{\actstart, a, id}$, with $a \in A$ and $id \in \mathbb{N}$, indicating the time at which an instance of $a$ identified with $id$ is started;
  $\tuple{\actend, a, id}$, with $a \in A$ and $id \in \mathbb{N}$, indicating the time at which an instance of $a$ identified with $id$ is terminated;
  $\tuple{\condstart, c}$, with $c \in P$, indicating the time at which a durative condition starts;
  $\tuple{\condend, c}$, with $c \in P$, indicating the time at which a durative condition ends;
  $\tuple{\condinst, c}$, with $c \in P$, indicating the time at which an instantaneous condition is checked;
  $\tuple{\effadd, p}$, with $p \in P$, indicating the time at which an add effect takes place;
  $\tuple{\effdel, p}$, with $p \in P$, indicating the time at which a delete effect takes place.
  We write $kind(t)$ for the first element of a time-point $t$.
\end{definition}

\begin{definition}
  \label{def:cnd-time-points}
  The \deftitle{timed time-points of a set of add (resp. delete) effects} $E^+$ are $ttp(E^+) \eqbydef \{\tuple{\tau, \mktp{\effadd}{p}} \!\mid\! \tuple{\tau, p} \!\in\! E^+\}$ (resp. $ttp(E^-) \!\eqbydef\! \{\tuple{\tau, \mktp{\effadd}{p}} \!\mid\! \tuple{\tau, p} \!\in\! E^+\}$).
\end{definition}

\begin{definition}
  \label{def:eff-time-points}
  The \deftitle{timed time-points of a set of conditions} $C$ are a set $ttp(C)$ defined as $\{\tuple{\tau, \mktp{\condinst}{p}} \mid \tuple{[\tau, \tau], p} \in C\} \cup \{\tuple{\tau_1, \mktp{\condstart}{p}}, \tuple{\tau_2, \mktp{\condend}{p}} \mid \tuple{[\tau_1, \tau_2], p} \in C, \tau_1 \not = \tau_2\}$.
\end{definition}
\noindent
Given an action $a \in A$, we define the set of timed time-points of $a$ as $ttp(a) \eqbydef ttp(C_a) \cup ttp(E^+_a) \cup ttp(E^-_a)$. Similarly, we define the set of goals and tils timed time-points as $ttp(T,G) \eqbydef ttp(T^+) \cup ttp(T^-) \cup ttp(G)$.

Without loss of generality, from here on, we assume that for each
action $a$ in the ICE planning problem and for each duration, the
relative ordering of time-points in $ttp(a)$ is fixed for every $d \in
\sprationals$ such that $d^{min}_a \le d \le d^{max}_a$. It is easy to
see that if this is not the case, we can split the interval
$[d^{min}_a, d^{max}_a]$ in sub-intervals having such a property and
create a copy of action $a$ with appropriate duration constraints for
each sub-interval. The number of actions created by this
transformation is at most quadratic, because having $n$ timings as
$\rstart + k$ and $m$ specified as $\rend - k$, we can construct all
the possible total orderings respecting two total orderings of size
$n$ and $m$, that are at most $m \times n$.

The search proceeds by either starting new action instances, thus
adding new time-points (corresponding to the just-started action
events) in a ``todo-list'', or by consuming such time-points by
applying their effects and checking their conditions on the state. In
this way, we construct a total order of time-points that is
causally-valid. In order to symbolically maintain and check the
temporal constraints we use a Simple Temporal Network (STN)
\cite{stn}.

\begin{definition}
  \label{def:search-state}
  A \textbf{search state} is a tuple $\tuple{\mu, \delta, \sstodo, \sstn, \sslast}$ s.t.:
  \begin{itemize}
  \item $\mu \subseteq P$ records the predicates that are true in the state;
  \item $\delta$ is a multiset of predicates in $P$, representing the active durative conditions to be maintained valid;
  \item $\sstodo$ is a list of lists of time-points. It constitutes
    the ``agenda'' of future commitments to be resolved. $\sstodo$
    contains a list for TILs and goals having a timing relative to the
    plan start, one for those that are timed relative to the plan end,
    and a list for each action that has been started. Each list
    contains the time-points that have not been explored by the search
    yet. Crucially, all the lists are sorted according to the total
    order of time-points.
  \item $\sstn$ is an STN defined over time-points that stores and checks the metric and precedence temporal constraints;
  \item $\sslast$ is the last time-point evaluated in this search branch.
  \end{itemize}
\end{definition}

The initial state for our problem is defined according to
\cref{alg:initial-state}. Intuitively, we start from the propositional
initial state $I$, enriched with an STN in which the time-points
corresponding to all the TILs in $T$ and the goals in $G$ are prepared
and constrained to their prescribed timings. In order to enforce such
constraints, we create two special time-points, $\mkptp{\planstart}$
and $\mkptp{\planend}$ representing the plan execution beginning and
ending moments, respectively. All the constraints for TILs and goals
are expressed relatively to these time-points by directly translating
the ICE effects and conditions expressions as STN constraints. All the
time-points being created are collected in two sets $sl$ and $el$ that
contain the time-points whose timing is relative to the start of the
plan and to the end of the plan, respectively. These sets are used to
initialize the $\sstodo$ agenda with two ordered lists that contain
the time-points to be expanded. The reason why we use two lists is
because we have a direct total order between the time-points that are
relative to $\rstart$, and another total ordering for the time-points
constrained with $\rend$, but we do not know how these time-points are
interleaved. By using two lists we basically force the search to
non-deterministically select one of the top elements in $\sstodo$ to
expand.

\begin{algorithm}[tb]
\caption{Initial state computation}
\label{alg:initial-state}
\begin{algorithmic}[1]
  \Procedure{FillTN}{$t_\vdash$, $t_\dashv$, $E^+$, $E^-$, $C$, $\sstn$}
    \State{$sl, el \gets \emptyset, \emptyset$}
    \ForAll{$\tuple{\tau, t} \in ttp(C) \cup ttp(E^+) \cup ttp(E^-)$}
      \If{$\tau = \rstart + k$}
        \State{$sl \gets sl \cup \{\tuple{k, t}\}$}
        \State{\Call{PushTNConstraint}{$\sstn$, $t - t_\vdash = k$}}
      \ElsIf{$\tau = \rend - k$}
        \State{$el \gets el \cup \{\tuple{k, t}\}$}
        \State{\Call{PushTNConstraint}{$\sstn$, $t_\dashv - t = k$}}
      \EndIf
    \EndFor
    \State{\Return{$\tuple{sl, el}$}}
  \EndProcedure

  \Procedure{GetInit}{ }
    \State{$\sstn \gets $ \Call{MakeEmptyTN}{\ }}
    \State{$sl, el \gets $ \Call{FillTN}{$\mkptp{\planstart}$, $\mkptp{\planend}$, $T^+$, $T^-$, $G$, $\sstn$}}
    \State{$\sstodo \gets [$ \Call{SortByAscendingTime}{sl}, \Call{SortByDescendingTime}{el} $]$}
    \State{\Return{$\tuple{I, \emptyset, \sstodo, \sstn, \mkptp{\planstart}}$}}
  \EndProcedure
\end{algorithmic}
\end{algorithm}

Given a state $s$ as per \cref{def:search-state}, we define the set of
successors $\textsc{Succ}(s)$ as the following set of states:
$$
\scriptsize \{ \textsc{SuccTp}(s, tp) \mid \tuple{tp, \cdots} \in s.\sstodo\} \cup \{ \textsc{SuccAct}(s,a) \mid a \in A\}
$$
We have two kind of successors, namely the ones deriving from the
evaluation of time-points in $\sstodo$, by selecting one list and
expanding its head, and the ones obtained by starting new actions
instances. The expansion of the selected time-point $tp$ by means of
the \textsc{SuccTp} function in \cref{alg:succ-tp} is obtained by
either applying the effects (if $tp$ represents an effect, lines 4-7)
or by checking the conditions (if $tp$ represents conditions, lines
8-11) and by imposing the appropriate temporal constraints in the
STN. Moreover, if $tp$ is the start of a durative condition, we add
the predicate to be maintained to the $\delta$ multiset, so that no
effects violating the condition can be applied (because of line 7)
until the time-point ending the durative condition is expanded,
removing the condition from $\delta$ (line 11). Note that $\delta$ is
a multiset so that if two durative conditions on the same predicate
are active at the same time, we do not remove the condition upon the
termination of the first interval. The temporal constraints added to
the STN impose a total ordering (allowing contemporary nodes) on the
expanded time-points (line 12) and ``push'' the non-expanded
time-points to happen later with respect to the expanded
time-point. This constraints on the future commitments can be either
strict, (i.e. $>$) to impose a positive-time-separation between
time-points that are mutex (e.g. between an effect and a supporting
condition), or weak (i.e. $\ge$) to allow non-interfering effects or
multiple conditions to happen at the same time. Note that the STN is
checked upon expansion and, if found infeasible, the expansion is
aborted by returning $\emptyset$, signaling that the expansion of $tp$
is a dead-end.

\begin{algorithm}[tb]
\caption{Existing time-point expansion}
\label{alg:succ-tp}
\begin{algorithmic}[1]
  \Procedure{SuccTp}{$s$, $tp$}
    \State{$\tuple{\mu, \delta, \sstodo, \sstn, \sslast} \gets $ \Call{CopyState}{$s$}}
    \State{$\sstodo \gets $\Call{RemoveTP}{$\sstodo$, $tp$}} \Comment{This also pops empty lists from $\sstodo$}
    \If{$kind(tp) \in \{\effadd, \effdel\}$}
      \If{$tp = \tuple{\effadd, p}$}
        {$\mu \gets \mu \cup \{p\}$}
      \ElsIf{$tp = \tuple{\effdel, p}$}
        {$\mu \gets \mu / \{p\}$}
      \EndIf
      \If{$(\bigcup_{c \in \delta} c) \not \subseteq \mu$} \Return{$\emptyset$} \EndIf
    \ElsIf{$kind(tp) \in \{\condstart, \condend, \condinst\}$}
      \If{$c \not \in \mu$} \Return{$\emptyset$} \EndIf
      \If{$tp = \tuple{\condstart, c}$}
        {\Call{AddToMultiSet}{$\delta$, $c$}}
      \ElsIf{$tp = \tuple{\condend, c}$}
        {\Call{PopFromMultiSet}{$\delta$, $c$}}
      \EndIf
    \EndIf
    \State{\Call{PushTNConstraint}{$\sstn$, $tp \ge \sslast$}}
    \ForAll{$t \in l \mid l \in \sstodo$}
      \If{$tp = \tuple{\effadd, p} \vee tp = \tuple{\effdel, p}$}
        \If{$t = \tuple{\effadd, q} \vee t = \tuple{\effdel, q}$}
          \If{$p \not = q$}
            {\Call{PushTNConstraint}{$\sstn$, $t \ge tp$}}
          \Else
            { \Call{PushTNConstraint}{$\sstn$, $t > tp$}}
          \EndIf
        \ElsIf{$kind(t) \in \{\condstart, \condend, \condinst\}$}
          \State{\Call{PushTNConstraint}{$\sstn$, $t > tp$}}
        \Else
          { \Call{PushTNConstraint}{$\sstn$, $t \ge tp$}}
        \EndIf
      \Else
        { \Call{PushTNConstraint}{$\sstn$, $t \ge tp$}}
      \EndIf
    \EndFor
    \If{$\neg$ \Call{CheckTN}{$\sstn$}} \Return{$\emptyset$} \EndIf
    \State{\Return{$\{ \tuple{\mu, \delta, \sstodo, \sstn, tp} \}$}}
  \EndProcedure
\end{algorithmic}
\end{algorithm}

The second kind of expansion is the ``opening'' of new action
instances (\cref{alg:succ-act}), that is, we decide to start a new
action. This choice adds to $\sstodo$ a new list of
time-points ordered according to their constraints,
%
%
representing the commitments on the future that the action brings. The
computation of these commitments is analogous to the one described in
\cref{alg:initial-state} with the notable difference that $\rstart$
(resp. $\rend$) is interpreted against the time-point $t_{a\vdash}$
(resp. $t_{a\dashv}$) representing the beginning (resp. the ending) of
the action. Note that we assign a fresh instance id to $t_{a\vdash}$
and $t_{a\dashv}$ to maintain a correlation between an instance of an
action starting and its ending; this will be exploited in order to
reconstruct the plan to correlate time-points belonging to the same
action instance. Also in this case, we check the consistency of the
STN before confirming the successor to ensure the temporal feasibility
of the action-opening choice.

\begin{algorithm}[tb]
\caption{Action opening expansion}
\label{alg:succ-act}
\begin{algorithmic}[1]
  \Procedure{SuccAct}{$s$, $a$}
  \State{$\tuple{\mu, \delta, \sstodo, \sstn, \sslast} \gets $ \Call{CopyState}{$s$}}
    \State{$id \gets $ \Call{MkFreshInstanceID}{\ }}
    \State{$t_{a\vdash}, t_{a\dashv} \gets \mkidtp{id}{\actstart}{a}, \mkidtp{id}{\actend}{a}$}
    \State{\Call{PushTNConstraint}{$\sstn$, $d^{min}_a \le t_{a\dashv} - t_{a\vdash} \le d^{max}_a$}}
    \State{$sl, el \gets $ \Call{FillTN}{$t_{a\vdash}$, $t_{a\dashv}$, $E^+_a$, $E^-_a$, $C_a$, $\sstn$}}
    \State{$\sstodo \gets \sstodo + [$ \Call{SortByActionTotalOrder}{$sl \cup el$, $a$} $]$}
    \If{$kind(l) \in \{\effadd, \effdel\}$}
      \State{\Call{PushTNConstraint}{$\sstn$, $t_{a\vdash} - l > 0$}}
    \Else
      { \Call{PushTNConstraint}{$\sstn$, $t_{a\vdash} - l \ge 0$}}
    \EndIf
    \ForAll{$t \in \sstodo$}
      {\Call{PushTNConstraint}{$\sstn$, $t - t_{a\vdash} \ge 0$}}
    \EndFor
    \If{$\neg$ \Call{CheckTN}{$\sstn$}} \Return{$\emptyset$} \EndIf
    \State{\Return{$\{ \tuple{\mu, \delta, \sstodo, \sstn, t_{a\vdash}} \}$}}
  \EndProcedure
\end{algorithmic}
\end{algorithm}

A goal state $\tuple{\mu, \delta, \sstodo, \sstn, \sslast}$ for our search
schema is a state where $\sstodo$ is empty, signaling that no
commitments on the future are still to be achieved. Note that goals
are automatically satisfied in such a state, because they are added to
$\sstodo$ in the initial state
(\cref{alg:initial-state}). Constructing a plan from a goal state can
be done by simply extracting a consistent model from $\sstn$ (such a
model is guaranteed to exists because the STNs are kept consistent by
the \textsc{SuccTp} and \textsc{SuccAct} successor functions) as
follows. Let $\beta : \mathcal{T} \rightarrow \prationals$ be a
consistent model for $\sstn$, where $\mathcal{T}$ is the set of the
time-points in $\sstn$. A solution plan encoded by $\beta$ is as follows.
$$
\{\tuple{\beta(s), a, \beta(e)-\beta(s)} \mid s = \mkidtp{i}{\actstart}{a}, e = \mkidtp{i}{\actend}{a}\}
$$
Intuitively, we take all the action startings ($s$) and the
corresponding endings ($e$) and create an action instance
in the plan that starts and lasts according to the STN model.

Computing subsumption of temporal states can be very hard
\cite{temporal-subsumption}, we employ a best-first tree-search
algorithm, using an $A^*$-like heuristic schema that sums the cost of
the path to a state $s$ (i.e. $g(s)$) with the heuristic estimation to
reach the goal $h(s)$; in the next section we will detail how $h(s)$
is computed in our framework.

\begin{algorithm}[tb]
\caption{Search algorithm}
\label{alg:search}
\begin{algorithmic}[1]
  \Procedure{Search}{\ }
    \State{$i \gets $ \Call{GetInit}{\ }; \ \ $ g(i) \gets 0$}
    \State{$Q \gets $ \Call{NewPriorityQueue}{\ }}
    \State{\Call{Push}{$Q$, $i$, $h(i)$}}
    \While{$c \gets $ \Call{PopMin}{$Q$}}
      \If{$|c.\sstodo| = 0$}
        {\Return{\Call{GetPlan}{$c.\sstn$}}}
      \Else
        \ForAll{$s \in $ \Call{Succ}{$s$}}
          \State{$g(s) \gets g(c) + 1$}
          \State{\Call{Push}{$Q$, $s$, $g(s) + h(s)$}}
        \EndFor
      \EndIf
    \EndWhile
  \EndProcedure
\end{algorithmic}
\end{algorithm}

\begin{theorem}
  Let $\mathcal{P} \eqbydef \tuple{P, A, I, T, G}$ be a planning
  problem with ICE, if $\mathcal{P}$ admits a solution plan, the
  search \cref{alg:search} terminates with a valid plan $\pi$.
\end{theorem}

Note that it is possible to soundly search for a plan by collapsing
states that have the same $\mu$ and the same $\sstodo$ sizing, loosing
the guarantee to eventually find a plan if it exists. This
under-approximation of the problem can be used as a pre-processing
step and we employ it in \tamer.

\myparagraph{Heuristic and Relaxation}
To guide our search, we define a relaxation of the input
temporal planning model into a classical planning model, allowing the
use of any domain-independent heuristic designed for classical
planning in our context.
The overall idea behind our relaxation is to create a classical
planning action for each time-point in the planning problem with ICE,
impose ordering constraints among time-points that we know are
ordered, and use such a relaxation to compute heuristic values, by
translating an arbitrary search state for the original ICE planning
problem into a classical state for the relaxation.

Our relaxation is defined as a STRIPS classical planning problem on a set of predicates $P'$ defined as $P \cup \{q^a_i \mid i \in \{0, \cdots, |ttp(a)|\}, a \in A\} \cup \{g_x \mid x \in ttp(T, G)\}$, having initial state $I \cup \{q^a_0 \mid a \in A\}$. We create a classical planning action for each time-point of each action in $A$, one for each TIL in $T$ and one for each goal in $G$. \Cref{alg:relaxed} formalizes the construction of  the set of relaxed classical actions $A'$: we characterize each action $a' \in A'$ with its precondition ($pre_{a'}$), add and delete effects ($\effadd_{a'}$ and $\effdel_{a'}$).
The additional predicates $q^a_i$ are used to impose precedences between time-points belonging to the same action. Initially each $q^a_0$ is true, signaling that no action is started; when an action starts executing its time-points, the ``true value'' is moved from $q^a_i$ to $q^a_{i+1}$ (i.e. $q^a_i$ is set to false and $q^a_{i+1}$ to true), until the last time-point is executed, at which point we reset $q^a_0$ to true. This is in fact a unary counter over the time-points of a durative action. The additional predicates $g_x$, instead, are used as the relaxation goals together with the $q^a_0$ predicates for each $a \in A$ (i.e. $G' \eqbydef \{g_x \mid g_x \in P\} \cup \{q^a_0 \mid a \in A\}$). The $g_x$ predicates are set to true by relaxed actions that represent a goal or a TIL, imposing to the relaxation that all the goals and TILs need to be encountered in a plan. For the sake of simplicity, we do not impose precedences among actions representing goals or TILs because these are not guaranteed to be totally-ordered.

\begin{algorithm}[tb]
\caption{Relaxed actions}
\label{alg:relaxed}
\begin{algorithmic}[1]
  \Procedure{GetRelaxationActions}{\ }
    \ForAll{$a \in A$}
      \State{$i \gets 0$}
      \ForAll{$\mktp{x}{p} \in $ \Call{SortByActionTotalOrder}{$ttp(a)$}}
        \State{$c, e^+, e^- \gets \emptyset, \emptyset, \emptyset$}
        \State{$j \gets (i+1) \bmod |ttp(a)|$}
        \If{$x = \effadd$}
          {$e^+ \gets \{p\}$}
        \ElsIf{$x = \effdel$}
          {$e^- \gets \{p\}$}
        \Else
          { $c \gets \{p\}$}
        \EndIf
        \State{$pre_{a'}, \effadd_{a'}, \effdel_{a'} \gets \{q^a_i\} \cup c, \{q^a_j\} \cup e^+, \{q^a_i\} \cup e^-$}
        \State{$A' \gets A' \cup {a'}; \:\: i \gets j$}
      \EndFor
    \EndFor
    \ForAll{$\mktp{x}{p} \in ttp(T, G)$}
      \State{$c, e^+, e^- \gets \emptyset, \emptyset, \emptyset$}
      \If{$x = \effadd$}
        {$e^+ \gets \{p\}$}
      \ElsIf{$x = \effdel$}
        {$e^- \gets \{p\}$}
      \Else
        { $c \gets \{p\}$}
      \EndIf
      \State{$pre_{a'}, \effadd_{a'}, \effdel_{a'} \gets c, \{g_{\mktp{x}{p}}\} \cup e^+, e^-$}
      \State{$A' \gets A' \cup {a'}$}
    \EndFor
    \State{\Return{A'}}
  \EndProcedure
\end{algorithmic}
\end{algorithm}
From the relaxed model, we can compute heuristic values as follows.
A given search state $s \eqbydef \tuple{\mu, \delta, \sstodo, \sstn,
  \sslast}$ corresponds to a state $s'$ defined as $\mu \cup \{q^a_0 \mid
a \in A\} \cup \{q^a_{|ttp(a)| - |l|} \mid l \in \sstodo,
\mkidtp{i}{\actend}{a} \in l\} \cup \{g_x \mid x \not \in \sstodo \wedge x
\in ttp(T, G)\}$ in the relaxed
model. Therefore, we can compute the heuristic value for $s$ as
$h(s')$, being $h$ any classical planning heuristic.

\myparagraph{Simultaneity Optimization}
The search schema above can be optimized by ``compressing'' multiple
time-points that happen at the same time in a single search step.  In
particular, it is possible to recognize multiple conditions and
multiple effects that must happen at the same time and force the
search to expand these nodes in a static, pre-fixed order without
intermediate branching.  We syntactically recognize three patterns of
such \emph{simultaneous} time-points.  The first are the effect
time-points (independently of whether they are add or delete)
belonging to the same action that are scheduled at the same time (e.g
two effects, both at $\rstart + k$).  The second are condition
time-points ($\condstart$, $\condend$, or $\condinst$) belonging to
the same action that are scheduled at the same time (e.g two
conditions, both at $\rstart + k$).  The third are time-points of
heterogeneous type (i.e. a condition and an effect) belonging to the
same action that are scheduled at the same time. For this last third
case, we need to be careful in checking that no other pair of
condition and effect belonging to a different action would require to
be executed simultaneously to the pair that we are compressing.

\section{Domain-Dependent Planner Generation}
\label{sec:compilation}

One of the major areas in which planning problems with ICE arise is
the orchestration of flexible production. In fact, it is not uncommon
to have some operation requiring a certain amount of time that must
immediately (or within some deadline) be followed by other
operations. Being this the intended deployment setting of our planner,
we recognize the existence of a ``deployment phase'' that is a moment
in which an instance of the planner is given a certain domain and it
will be required to solve lots of different problems on that
domain. For this reason, we equipped our planner with the capability
of generating compilable source code that embeds the search strategy
described in the previous section. In this way, we can greatly
simplify the computational work of the planner implementation by
``hardcoding'' and optimizing parts of the problem such as the number
of objects and the actions, obtaining performances that get closer to
domain-dependent planners implementations.

\begin{figure}
  \resizebox{\columnwidth}{!}{\begin{tikzpicture}
  [
    plan/.style = {cylinder, shape border rotate=90, draw=black, minimum height=1.5cm, minimum width=1cm, shape aspect=.25, very thick, align=center, text width=1cm, fill=yellow!40},
    probs/.style = {cylinder, shape border rotate=90, draw=black, minimum height=1.5cm, minimum width=1cm, shape aspect=.25, very thick, align=center, text width=1cm, fill=green!20},
    data/.style = {cylinder, shape border rotate=90, draw=black, minimum width=2.5cm, shape aspect=.25, very thick, align=center, text width=2.5cm},
    tool/.style = {rectangle, draw=black, thick, minimum height=1.5cm, minimum width=2cm, very thick,  align=center, text width=2cm},
    link/.style = {-latex, very thick}
  ]

  \node[data, fill=blue!20] (model) {\small $\tuple{P, A, I, T, G}$ Planning Problem {\tiny (ANML/PDDL)}};
  \node[tool] (compiler) [below=1cm of model] {Planner Generation};
  \node[data, fill=yellow!40] (src) [right=1cm of compiler] {\small Domain-Dependent Planner Source {\tiny (C++)}};
  \node[tool] (gcc) [right=1cm of src] {C++ Compiler};
  \node[tool, fill=blue!20] (exe) [right=1cm of gcc] {Domain-Dependent Planner Executable};
  \node[data, fill=green!20] (instance) [above=1cm of exe] {\small $\tuple{P', A', I', T', G'}$ Planning Problem {\tiny (ANML/PDDL)}};
  \node[data, fill=red!20] (plan) [right=1cm of exe] {\ \\Plan\\\ };

  \draw[link] (model) -- (compiler);
  \draw[link] (compiler) -- (src);
  \draw[link] (src) -- (gcc);
  \draw[link] (gcc) -- (exe);
  \draw[link] (instance) -- (exe);
  \draw[link] (exe) -- (plan);
\end{tikzpicture}}
  \caption{\label{fig:compilation-flow} The compilation dataflow.}
\end{figure}
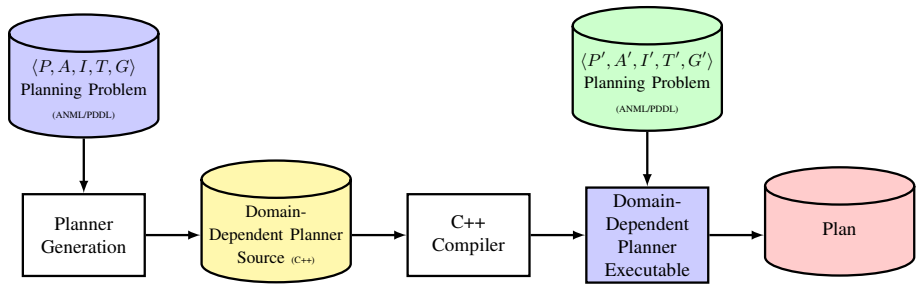

The dataflow of the compilation is shown in
\cref{fig:compilation-flow}. Starting from a planning problem
instance $\tuple{P,A,I,T,G}$, the generator produces a self-contained
C++ source code implementing the heuristic search described in the
previous section, fixing both the number of objects and the set of
actions of the problem. All the parts of the approach benefit from
this assumption: the state is implemented as a class that has one
field for each predicate and each time-point is given a unique ID to
simplify memory management and transforming all the needed maps into
arrays. The program obtained by compiling the generated source code
can solve the original problem or any other problem
$\tuple{P',A',I',T',G'}$ having a set of objects\footnote{While the
  formalization in the previous section assumed (as customary) a
  ground model, the code generation as well as the implementation
  accept a lifted model, where a certain number of object instances can
  be used to parametrize actions and predicates.}  that is a subset of
the objects of the original problem. In fact, we allow the user to
override the initial state and the goal of the problem as well as to
disable some of the objects in the original problem.

In situations where we have a family of problems having the same
structure and differing only because of the number of objects, initial
state and goals, we can use this compilation by creating a planning
problem that has at least as many objects as the biggest instance and
create a compiled executable with our technique. The very same
executable can then be used to solve all the instances in the batch
efficiently. Note that in this schema, the cost of compilation (that
is non-negligible for complex problem instances) is payed only once
for the entire set of instances sharing the same structure. This is
particularly useful in flexible production deployment, because domain
experts can safely estimate the maximum number of items, machines and
personnel for a specific factory/deployment and we can thus use this
information to compile (only once) an efficient domain-dependent
planner. Finally, note that our implementation is still capable of
symbolic planning (i.e. planning without using the intermediate
compilation step), granting a safe fall-back if for some reason one
instance is larger or different from the one used for compilation.

\section{Related Work}
\label{sec:rw}

Intermediate conditions and effects have been recognized as an
important feature for temporal planning in several works.
\cite{micheli-aij} indicates a practical case for ICE: the
authors propose a compilation from planning problems having temporal
uncertainty in the duration of PDDL 2.1 actions that produces plain
temporal planning problems with ICE.

In his commentary to the PDDL 2.1 language,
\cite{container-actions}
recognizes how the limitations imposed by the language make it
``exceptionally cumbersome'' to encode ICE. Another instance of
compilation of ICE into PDDL 2.1 can be found in \cite{clip-actions}
where the authors explain the so-called ``clip-action'' construction
that can be used to ``break'' a durative actions having ICE into a number
of smaller durative actions without ICE at the cost of adding required
concurrency \cite{cushing_temporally_expressive}, additional fluents
and additional (clip) actions.
These compilations can be used to transform a planning problem with
ICE into a PDDL 2.1 planning problem\footnote{Formally, this is only
  possible if we assume a $\epsilon$-separation semantics, but we
  disregard this detail for the rest of the paper.}, making it
possible to use any state of the art temporal planner for PDDL 2.1 to
solve problems with ICE. The current best planners for PDDL 2.1 are
based on heuristic search (e.g. \cite{popf,tfd}); in this paper we
build on these approaches and propose a technique that natively
supports and reasons on ICE. Our experiments indicate that this
approach is far superior to PDDL compilations.

The ANML planning language \cite{anml} directly and explicitly
supports ICE, in fact our planner, \tamer, is based on this
language. Unfortunately, few planners currently support ANML and none
of them implements a heuristic-search approach. In fact, \fape
\cite{fape} implements a plan-space search focusing on hierarchical
task decomposition (but still supports ICE with an integer time
interpretation), while \lcp \cite{lcp} implements an encoding of the
bounded planning problem into a constraint satisfaction problem.

A recent publication \cite{tpack} proposed the use of temporal metric
trajectory constraints to encode the temporal features of planning
problems. The authors focus on a language where all actions are
instantaneous, and ICE can be expressed by creating an action for
each time-point and imposing appropriate trajectory constraints. In
this paper, we retain the ``actions-as-intervals'' idea adopted by
both PDDL 2.1 and ANML, and experimentally show that this can give
advantages on some domains (in particular on one proposed by
\cite{tpack}).

Finally, we complement our contribution by presenting and evaluating a
compilation of our planner technique into executable code. We borrowed this idea
from the verification community, in particular from
the \spin model-checker \cite{spin}. In planning, this
idea has been recently exploited by \cite{geffner-simulated}
 where the planning actions are seen as
opaque simulators changing the planning state. The idea evolved in
the IW planner \cite{iw} that can solve black-box classical planning
problems by incrementally bounding the ``width'' of the problem. We
highlight that all these approaches are designed and work for
classical planning, while here we tackle temporal planning. Moreover,
we do retain domain-independent heuristic computation even in the
compiled executable as we start from a ``white-box'' model of the system.

\section{Experimental Evaluation}
\label{sec:experiments}

We experimentally evaluate the merits of both our search schema and
the impact of the code generation on a comprehensive set of benchmarks,
similar to the one used in \cite{tpack}. In particular, we took all
the \majsp, Temporal IPC and Uncertainty IPC instances (we disregarded
the ``\hsp'' instances that are not directly expressible in a planning
problem with ICE).
The \majsp domain is a job-shop scheduling problem in which a fleet of
moving agents transport items and products between operating
machines. We took all the 240 instances of this domain and we manually
re-coded them in ANML using ICE.
The Temporal IPC class is composed of temporal planning domains
(without ICE) of the IPC-14 competition
\cite{vallati20152014} for a total of 98 planning instances.
The Uncertainty IPC class consists of the same planning instances
where the durations of some actions are assumed to be uncontrollable
and the rewriting in \cite{micheli-aij} is used to produce
equivalent temporal planning problems with ICE.
Finally, we added a new domain, called \painter: a worker has to apply
several coats of paint on a set of items guaranteeing a minimum and a
maximum time between two subsequent coats on the same item. We created
300 instances of this domain by scaling the number of coats (from 2 to
11) and items (from 1 to 30) and formulated each instance in both
ANML, TPP (the language of \tpack) and PDDL 2.1.

We implemented the heuristic search technique of \cref{sec:search} in
a planner called \tamer. Our planner is written in C++ and accepts
both PDDL 2.1 and ANML specifications as input (recall that ANML
allows ICE, while PDDL 2.1 does not). \tamer uses the standard
$h_{add}$ classical planning heuristic \cite{hadd} on the relaxation
defined in \cref{sec:search}.
The domain-dependent generator of \cref{sec:compilation} is implemented
in C++ and we use the GCC C++ compiler to produce the executables from the generated code.

In our experimental analysis we compare against a number of
planners.
We use \itsat \cite{itsat} and \optic \cite{benton:12:optic} as
representatives of PDDL 2.1 planners: since PDDL 2.1 does not natively
support ICE, we employ the clip-construction described in
\cite{clip-actions} and the container-construction described in
\cite{container-actions} to capture the ICE semantics.
To use \itsat, we re-scaled the actions durations to integer values in
the Uncertainty IPC class (this is remarked by an asterisk in the
tables). We did the same for the \painter and \majsp domains, but
\itsat always crashed.
We also consider two ANML planners, namely \fape \cite{fape} and
\anmlsmt; the latter is an adaptation of the encoding described in
\cite{shin2005processes} for the ANML language\footnote{We implemented
  \anmlsmt using the MathSAT5 \cite{mathsat5} solver. The encoding is
  similar to \cite{lcp}.}. We could not use \fape for \majsp because
\fape does not support numerical fluents nor for the IPC domains
because \fape, differently from \anmlsmt, is unable to parse PDDL 2.1.
Finally, we include the \tpack planner \cite{tpack} in our
experiments, by manually recoding all the new instances in the \tpack
input language.

To evaluate the performance of the code-generation optimization, we
compiled the (automatically-generated) domain-dependent planner for
the largest instance of both \painter and \majsp, and used the
resulting executable to solve all the instances for that domain. The
GCC compilation times for such domains are 203 and 492 seconds,
respectively.  We indicate this approach as \comptamer.
We ran all the experiments on a Xeon E5-2620 2.10GHz with 1800s/15GB
time/memory limits.
%

\begin{figure*}
  \centering
  \begin{subfigure}[b]{.564\textwidth}%
    \resizebox{\textwidth}{!}{%
      \setlength\tabcolsep{3.8 pt}%
      \renewcommand{\arraystretch}{0.9}%
      \small%
      \begin{tabular}{|c|c|cccc||c|}
        \hline
        \multirow{8}{*}[-0.3cm]{\rotatebox{90}{\scriptsize Temporal IPC (a)}} & \scriptsize  Domain (\# inst.) &
        \itsatcol & \opticcol & \tpackcol & \anmlsmtcol & \tamercol \\[4pt]
        \cline{2-7}
        &&&&&&\\[-6pt]
        &  \scriptsize\Driverlog (20)   & \bf 18 & 15 &     14 & 4 & 12 \\
        &  \scriptsize\FloorTile (8)    & \bf  8 &  7 &      4 & 0 & 7 \\
        &  \scriptsize\MAP (20)         &     15 &  0 & \bf 20 & 0 & 8 \\
        &  \scriptsize\Matchcellar (10) & \bf 10 &  9 &      6 & 6 & 7 \\
        &  \scriptsize\Satellite (20)   & \bf 20 & 14 &      9 & 3 & 9 \\
        &  \scriptsize\TMS (20)         & \bf 20 &  1 &      1 & 0 & 0 \\
        \cline{2-7}
        &&&&&&\\[-6pt]
        &  \scriptsize Total (98)       & \bf 91 & 46 &     54 & 13 & 43 \\
        \hline
      \end{tabular}%
    }%
    \vskip 0.04cm%
    \resizebox{\textwidth}{!}{%
      \setlength\tabcolsep{3.5 pt}%
      \renewcommand{\arraystretch}{0.9}%
      \small%
      \begin{tabular}{|c|c|cccccc||c|}
        \hline
        \multirow{8}{*}[-0.3cm]{\rotatebox{90}{\scriptsize Uncertainty IPC (b)}} & \scriptsize Domain (\# inst.) & \itsatclipcol &  \itsatcontainercol & \opticclipcol &  \opticcontainercol &  \tpackcol & \anmlsmtcol & \tamercol \\
        \cline{2-9}
        &&&&&&&&\\[-6pt]
        & \scriptsize\Driverlog (20)   &     0 & 0 & 0 &      5 & \bf 7 &  4 &     5 \\
        & \scriptsize\FloorTile (8)    &     0 & 0 & 0 &      0 & \bf 4 &  0 & \bf 4 \\
        & \scriptsize\MAP (20)         &     0 & 0 & 0 &      0 & \bf 2 &  0 &     1 \\
        & \scriptsize\Matchcellar (10) &     4 & 0 & 3 & \bf 10 &     5 &  5 &     5 \\
        & \scriptsize\Satellite (20)   &     1 & 0 & 0 &      1 & \bf 5 &  1 &     1 \\
        &  \scriptsize\TMS (20)        & \bf 1 & 0 & 0 &      0 &     0 &  0 &     0 \\
        \cline{2-9}
        &&&&&&&&\\[-6pt]
        & Total (98)                   &     7 & 0 & 3 &     16 & \bf 23 & 10 &    16 \\
        \hline
      \end{tabular}%
    }
    \phantomsubcaption{\label{tbl:tempo-sat}\label{tbl:uncertainty}}
  \end{subfigure}
  \hfill
  \begin{subfigure}[b]{.400\textwidth}
    \resizebox{\textwidth}{!}{%
      \setlength\tabcolsep{3 pt}%
      \renewcommand{\arraystretch}{0.9}%
      \small%
      \begin{tabular}{|c|c|ccccc||c|}
        \hline
        \multirow{12}{*}[-0.15cm]{\rotatebox{90}{\scriptsize \painter (c)}} &\scriptsize \#Coats & \opticclipcol &  \opticcontainercol &  \tpackcol & \fapecol & \anmlsmtcol & \tamercol \\
        \cline{2-8}
        &&&&&&&\\[-6pt]
        &  2 & 0 & 3 & 2 & 2  & 1 & \bf 30 \\
        &  3 & 0 & 3 & 2 & 2  & 0 & \bf 30 \\
        &  4 & 3 & 4 & 7 & 11 & 6 & \bf 30 \\
        &  5 & 1 & 4 & 5 & 4  & 4 & \bf 30 \\
        &  6 & 0 & 4 & 4 & 3  & 3 & \bf 30 \\
        &  7 & 0 & 4 & 3 & 3  & 2 & \bf 30 \\
        &  8 & 0 & 3 & 2 & 2  & 2 & \bf 30 \\
        &  9 & 0 & 3 & 2 & 2  & 2 & \bf 30 \\
        & 10 & 0 & 3 & 2 & 2  & 2 & \bf 30 \\
        & 11 & 0 & 3 & 2 & 2  & 1 & \bf 30 \\
        \cline{2-8}
        &&&&&&&\\[-6pt]
        & Tot. & 4 & 34 & 31 & 33 & 23 & \bf 300 \\
        \hline
        \multicolumn{8}{c}{}\\[-6pt]
        \hline
        \multirow{6}{*}[-0.3cm]{\rotatebox{90}{\scriptsize \majsp (d)}} &\scriptsize \#Jobs\  & \opticclipcol &  \opticcontainercol &  \tpackcol & \fapecol & \anmlsmtcol & \tamercol \\
        \cline{2-8}
        &&&&&&&\\[-6pt]
        & 1 & 14 & 0 & 56 & NA & \bf 60 &     56 \\
        & 2 & 0  & 0 & 45 & NA &     45 & \bf 52 \\
        & 3 & 0  & 0 & 27 & NA &     30 & \bf 46 \\
        & 4 & 0  & 0 & 14 & NA &     17 & \bf 43 \\
        \cline{2-8}
        &&&&&&&\\[-6pt]
        & Tot. & 14 & 0 & 142 & NA & 152 & \bf 197 \\
        \hline
      \end{tabular}%
    }
    \phantomsubcaption{\label{tbl:painter}\label{tbl:majsp}}
  \end{subfigure}
  \caption{\label{fig:coverage}Coverage results for IPC-14 (a-b) and for \painter and \majsp (c-d).}
\end{figure*}

\begin{figure}[h]
    \resizebox{\textwidth}{!}{\input{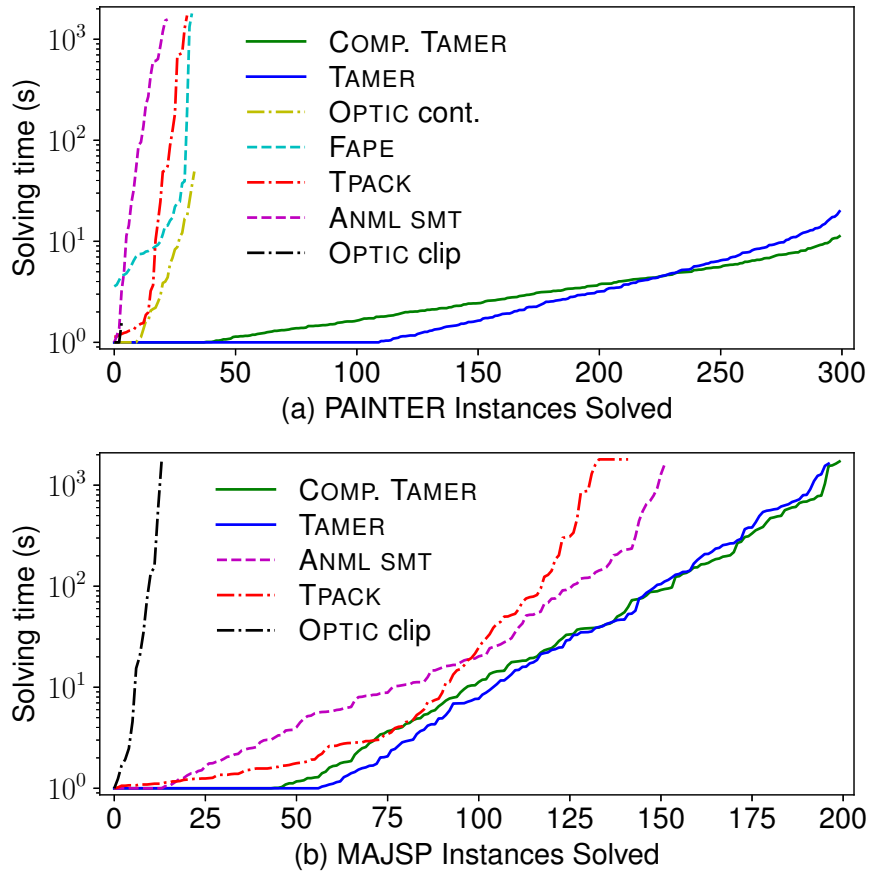}}
  \caption{\label{fig:plots}Cactus plots for the industrial domains (a-b).}
\end{figure}

\myparagraph{Results}
The coverage results for our experiments are presented in the tables
of \cref{fig:coverage}. Figures \ref{fig:plots}a-b depict the time
performance on the \painter and \majsp domains, respectively.

The Temporal IPC domains show how, on domains without ICE, our planner
exhibits comparable, but inferior, performance w.r.t. \optic and
\tpack. This is expected, as our search schema introduces some
overhead in order to support ICE. Also, IPC domains are very
challenging from a ``classical planning'' point of view, but are not
very rich in terms of temporal features. This fact is reflected also
in \cref{fig:coverage}b, where our planner is able to solve the second
highest number of instances behind \tpack (\optic achieves the same
coverage, mostly due to MatchCellar, where it excels). In fact, the
Uncertainty IPC instances have very simple and localized ICE, and the
propositional complexity is predominant.

\Cref{tbl:painter} shows that \tamer outperforms all the competitors
in all the \painter instances. In fact, we tried to further scale the same
domain (up to 32 coats and 30 items) and \tamer is still
able to solve all the instances. We highlight how all the
competitors (excepting Optic with the clip-action construction) solve
almost the same number of instances, that are an order of magnitude
less that those solved by \tamer. This is also evident from the cactus
plot in \cref{fig:plots}a that also clearly depicts how the
generated domain-dependent planner (produced from the
largest ANML model by our approach) surpasses the performance of the
symbolic search as the size of the instances grows. In this schema,
the GCC compilation cost is payed only once and the resulting solver is
used uniformly on all the instances. The advantage of the generated
domain-dependent planner is vast: note that the time scale of the plot
is logarithmic and that the domain-dependent planner cuts the run-time in half
for the harder instances.
The situation is similar in the \majsp domain. We highlight that these
instances are taken from the \tpack distribution and both \tamer and
\anmlsmt are able to outperform \tpack. In this case, the impact of
the code generation is less dramatic, but \comptamer is able to
solve 200 instances in total, beating any other approach.

Finally, we remark that our code generation procedure can be invoked
on each problem instance, producing executables that are tailored to
a specific set of objects but allow the change of the initial
state and/or the goal. The GCC compilation time is
non-negligible for larger instances, but the run-time is always
smaller than \tamer: the average speedup is 243\% and we
managed to solve 205 instances of \majsp (still accounting the compilation
time in the timeout).

\section{Conclusions}
\label{sec:conclusion}

In this paper, we presented a novel heuristic-search technique for
solving planning problems exhibiting intermediate conditions and
effects (ICE), that are useful to naturally model and effectively
solve practical and industrial problems. The technique is complemented
by code generation capabilities that further push the performance of
the solver.

Future work includes the exploration of different heuristics on our
relaxation of the problem and the definition of alternative
relaxations. Moreover, considering optimality for problems with ICE is
another interesting research line.

\bibliographystyle{plain}
\bibliography{refs.bib}

\end{document}